\documentclass[letterpaper]{article} 
\usepackage{aaai2026}  
\usepackage{times}  
\usepackage{helvet}  
\usepackage{courier}  
\usepackage[hyphens]{url}  
\usepackage{graphicx} 
\usepackage{natbib}  
\usepackage{caption} 
\usepackage{algorithm}
\usepackage{lineno}
\usepackage{algpseudocode}
\usepackage{amsmath}
\usepackage{array}
\usepackage{booktabs}
\usepackage{amsmath} 
\usepackage{multirow}
\usepackage{newfloat}
\usepackage{listings}
\usepackage{{algorithm}}
\DeclareCaptionStyle{ruled}{labelfont=normalfont,labelsep=colon,strut=off} 
\floatstyle{ruled}
\newfloat{listing}{tb}{lst}{}
\floatname{listing}{Listing}
\title{GEM: Gaussian Embedding Modeling for \\Out-of-Distribution Detection in GUI Agents}
\author{
Zheng Wu, Pengzhou Cheng, Zongru Wu, Lingzhong Dong, Zhuosheng Zhang\thanks{Corresponding author. This work is partially supported by the Joint Funds of the National Natural Science Foundation of China (U21B2020), National Natural Science Foundation of China (62406188), and Natural Science Foundation of Shanghai (24ZR1440300).}
}
\affiliations{
    School of Computer Science, Shanghai Jiao Tong University\\
    \{wzh815918208, zhangzs\}@sjtu.edu.cn
}

\usepackage{bibentry}

\begin{document}

\maketitle

\begin{abstract}
Graphical user interface (GUI) agents have recently emerged as an intriguing paradigm for human-computer interaction, capable of automatically executing user instructions to operate intelligent terminal devices. 
However, when encountering out-of-distribution (OOD) instructions that violate environmental constraints or exceed the current capabilities of agents, GUI agents may suffer task breakdowns or even pose security threats. Therefore, effective OOD detection for GUI agents is essential. Traditional OOD detection methods perform suboptimally in this domain due to the complex embedding space and evolving GUI environments. 
In this work, we observe that the in-distribution input semantic space of GUI agents exhibits a clustering pattern with respect to the distance from the centroid.
Based on the finding, we propose GEM, a novel method based on fitting a Gaussian mixture model over input embedding distances extracted from the GUI agent that reflect its capability boundary. 
Evaluated on 8 datasets spanning smartphones, computers, and web browsers, our method achieves an average accuracy improvement of 23.70\% over the best-performing baseline while only increasing training time by 4.9\% and testing time by 6.5\%.
We also experimentally demonstrate that GEM can improve the step-wise success rate by 9.40\% by requesting assistance from the cloud model when encountering OOD samples.
Analysis verifies the generalization ability of our method through experiments on nine different backbones. 

\end{abstract}

\begin{links}
\link{Code}{https://github.com/Wuzheng02/GEM-OODforGUIagents}
\end{links}

\begin{figure*}[t]
  \centering
  \includegraphics[width=0.9\textwidth]{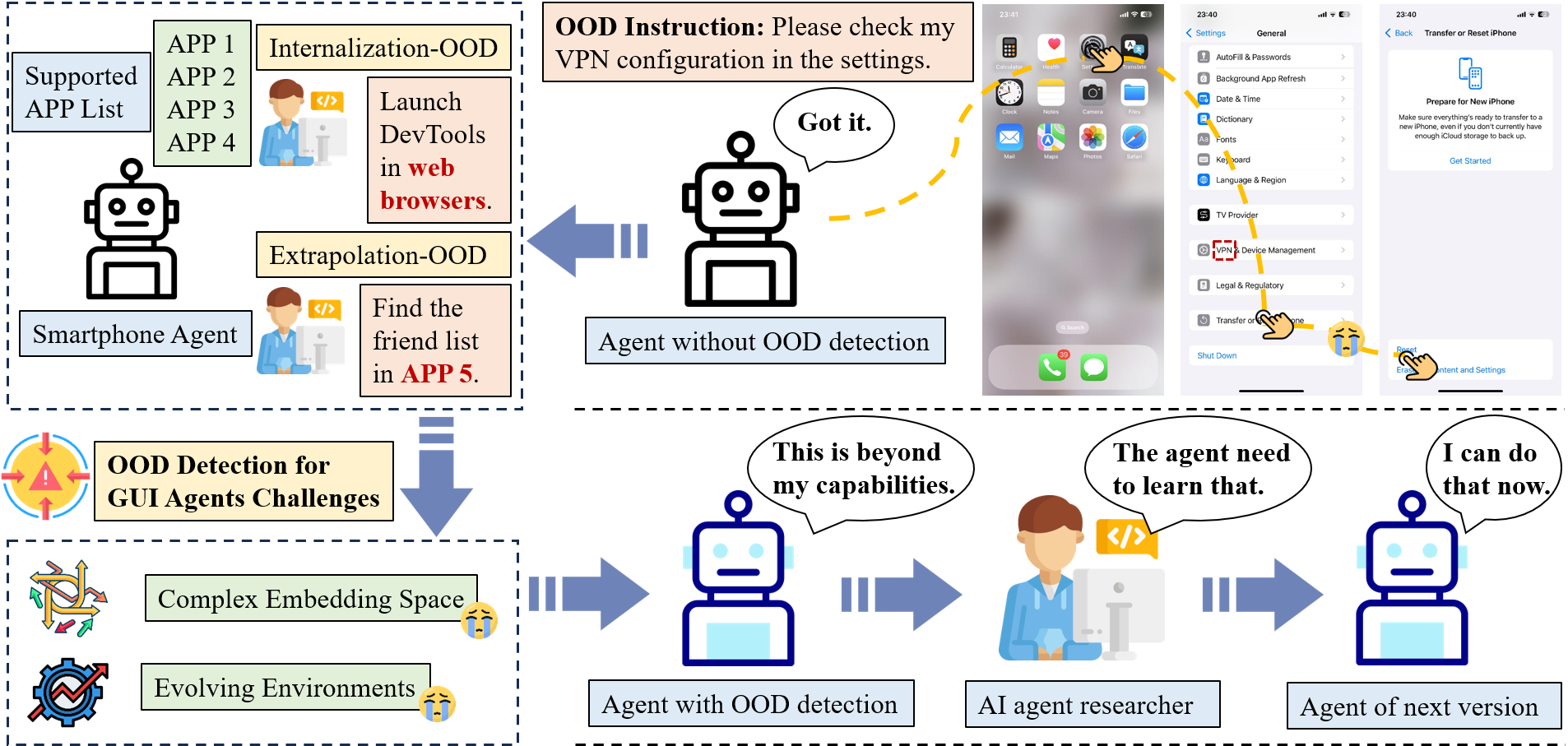}
  \caption{Comparison between an agent w/ and w/o OOD detection when facing an OOD instruction. Also illustrated are the OOD scenarios and the challenges of OOD detection for GUI agents.}
  \label{intro1}
\end{figure*}
\section{Introduction}
Recently, graphical user interface (GUI) agents~\citep{zhang2024large,tang2025surveymllmbasedguiagents} have emerged as an intriguing paradigm to human-computer interaction, capable of autonomously executing user instructions and performing human-like control on intelligent terminal devices such as smartphones, computers, and web browsers. 
The common approach to building GUI agents involves post-training multimodal large language models (MLLMs) using high-quality trajectory data to enhance key task capabilities such as perception~\cite{li2023blip,wang2023image}, reasoning~\cite{yao2022react,wei2022chain}, and reflection~\citep{liu2025infiguiagent,hu2025guiding}.
Despite notable advances in instruction following, GUI agents remain vulnerable to out-of-distribution (OOD) risks---executing instructions that violate environmental constraints (e.g., non-existent functions or applications) or exceed the agent’s current capabilities. 
These failures can lead to task breakdowns or even pose security threats.
In real-world applications, OOD risks for GUI agents come with two primary forms:
(i) \textbf{Internalization-OOD}, where the agent operates in domain-specific environments (e.g., a particular type of smartphones or vehicle cabins) but incorrectly assumes the presence of unsupported capabilities;
(ii) \textbf{Extrapolation-OOD}, where the agent encounters instructions tied to dynamic or evolving environments (e.g., new third-party applications). 

As illustrated in Figure~\ref{intro1}, following OOD instructions and executing an incorrect action path (e.g., unintentionally resetting a smartphone) can result in critical failures. 
Therefore, it is essential for GUI agents to incorporate OOD detection mechanisms that identify tasks beyond their supported scope. This not only mitigates potential operational risks but also enables targeted enhancements to agent capabilities through continued research and development.

Popular OOD detection methods can be broadly categorized~\cite{malinin2018predictive} into two types: embedding-based approaches~\cite{lee2018simple} and model uncertainty-based approaches~\cite{hendrycks2016baseline}. 
These methods have demonstrated effectiveness in traditional (M)LLM tasks such as summarization~\cite{ren2022out}, visual question answering~\cite{kervadec2021roses}, mathematical reasoning~\citep{wang2024embedding}, and text generation~\cite{wu2022multi}. 
Directly applying these existing OOD detection techniques to the GUI agent domain results in suboptimal performance.
Notably, even with the classification threshold calibrated to correctly identify 95\% of in-distribution (ID) tasks, the best-performing method still misclassifies between 31.75\% and 48.92\% of OOD tasks as ID.

Compared with traditional (M)LLM-based OOD detection tasks, OOD detection for GUI agents presents two critical challenges:

\textbf{(i) Complex Embedding Space}: The inputs to GUI agents are inherently complex~\citep{ma2024comprehensive}. 
Concretely, GUI screens typically contain densely populated UI components~\cite{zhang2024large}, leading to higher information density than traditional MLLM tasks. 
Besides, the diversity of user instructions further complicates the inference of user intent. 
This complexity substantially increases the difficulty of effective OOD detection for GUI agents.

\textbf{(ii) Evolving Environments}: GUI environments frequently change due to system upgrades or the installation of new third-party applications.
As a result, GUI agents must continually adapt their capabilities~\citep{wang2025mobileE,liu2025learnact}. This evolving nature complicates OOD detection by demanding both accurate capability assessment and temporal adaptability. Consequently, reliance on static external classifiers becomes increasingly impractical.

To address the challenges above, we propose \textbf{GEM}, a novel method based on fitting a Gaussian mixture model (GMM)~\cite{reynolds2009gaussian} over input embedding distances extracted from the GUI Agent that reflect its capability boundary. 
Specifically, we first extract the input embeddings from the encoder layer of the GUI agent on its training data, and fit them into a high-dimensional hypersphere. 
We then compute the L2-norm distances of all embeddings relative to the centroid of the hypersphere. 
Under bayesian information criterion (BIC)~\cite{neath2012bayesian} supervision, we fit a GMM to these distances and define the OOD detection boundary as a configurable number of standard deviations away from each GMM cluster center.
Experiments across eight GUI agent datasets spanning smartphones, computers, and web browsers platforms show that our method consistently outperforms all traditional OOD detection methods. And we provide a possible method to use GEM to improve the end-to-end results for GUI agents.

In summary, we make three key contributions:

(i) We present the first systematic analysis of OOD detection for GUI agents and compare various widely adopted OOD detection methods in this domain.

(ii) We propose a novel GMM-based approach that leverages input embedding distances from MLLMs for OOD detection. Our method achieves an average accuracy improvement of 23.70\% over the best performing baseline across eight datasets spanning three platform types while only increasing training time by 4.9\% and testing time by 6.5\%.

(iii) We offer some new insights into this emerging research area and provide a possible method to use GEM to improve the end-to-end results for GUI agents.

\begin{table*}[t]
  \centering
\begin{tabular}{>{\centering\arraybackslash}m{3.6cm}
                >{\centering\arraybackslash}p{1.8cm}
                >{\centering\arraybackslash}p{1.8cm}
                >{\centering\arraybackslash}p{1.8cm}
                >{\centering\arraybackslash}p{1.8cm}
                >{\centering\arraybackslash}p{1.8cm}
                >{\centering\arraybackslash}p{1.8cm}}
\toprule
\multirow{2}{*}{\centering Method} 
    & \multicolumn{2}{c}{Smartphone} 
    & \multicolumn{2}{c}{Computer} 
    & \multicolumn{2}{c}{Web browser} \\
      \cmidrule(r){2-3} \cmidrule(r){4-5} \cmidrule(r){6-7}
      & AUROC $\uparrow$ & FPR95 $\downarrow$ & AUROC $\uparrow$ & FPR95 $\downarrow$ & AUROC $\uparrow$ & FPR95 $\downarrow$ \\
      \midrule
      \multicolumn{7}{c}{ \textbf{Embedding-based methods}} \\
      \midrule
      TV score     & 54.26 & 98.02 & 60.13 & 85.37 & 60.27 & 89.48 \\
      Last layer embedding  & 50.94 & 76.48 & 64.31 & 68.73 & 67.51 & 74.58\\
      Best layer embedding  & \textbf{76.14} & \textbf{48.92} & \textbf{89.72} & \textbf{45.60} & \textbf{89.77} & \textbf{31.75}\\
      \midrule
      \multicolumn{7}{c}{ \textbf{Uncertainty-based methods}} \\
      \midrule
      Top-k confidence   & 67.07 & 85.10 & 57.51 & 93.59 & 57.40 & 94.41 \\
      Output entropy  & 67.07 & 92.69 & 57.51 & 86.22 & 57.40 & 86.73 \\
      \bottomrule
    \end{tabular}
    \caption{Results of the pilot study. Existing popular OOD detection methods perform poorly in the GUI agent domain.}
  \label{pilot}
\end{table*}

\section{Related Work}
\subsection{MLLM-based GUI Agents}
With the advancement of MLLMs~\citep{liu2023visual,achiam2023gpt}, numerous GUI agent foundation models~\citep{hong2024cogagent,wu2024atlas,qin2025ui} have emerged, capable of executing user instructions across smartphones, computers, and web browsers. 
For the technical framework, prompt-based approaches~\cite{wang2025mobileE, jiang2025appagentx,wu2025quick} have been developed, leveraging closed-source models to construct agent systems that fulfill user instructions. 
Additionally, researchers have also explored pre-training~\cite{ma2024comprehensive,wu2025smoothing}, supervised fine-tuning (SFT)~\cite{zhang2023you,wu2025verios}, and reinforcement learning (RL)~\cite{zhoudigirl,xia2025gui,tang2025guig2gaussianrewardmodeling} techniques to further enhance the ability of GUI agents to complete user instructions. 
Regardless of their advancements, GUI agents inevitably face capability limitations. 
In complex and dynamic real-world environments~\cite{chengkairos,guo2025atomic}, GUI agents are prone to encountering OOD situations. 

\subsection{OOD Detection in MLLMs}
For OOD detection in MLLMs, it is necessary to jointly consider both visual and textual modalities, in contrast to traditional computer vision OOD detection, which only requires modeling the visual modality~\cite{liu2020energy,ren2019likelihood}, or LLMs OOD detection, which focuses on textual modality~\cite{wang2024embedding,ren2022out}. 
In the context of MLLMs, some researchers have proposed using maximum concept matching~\cite{ming2022delving} to characterize OOD uncertainty, while others have approached the problem by training models on ID datasets and classifying whether an input image belongs to an unknown category~\cite{wang2023clipn}. 
However, OOD detection for MLLMs remains a challenging research area~\cite{dong2024multiood,lu2024recent}. 
Furthermore, for MLLM-based GUI agents, task scenarios are significantly more complex and dynamic~\cite{shi2025towards} compared to traditional MLLM tasks, making OOD detection even more difficult.

\section{Investigating the Challenge of OOD Detection for GUI Agents}
\label{pilot_experiments}
In this section, we conduct pilot experiments to evaluate the effectiveness of popular OOD detection methods for GUI agents and understand the challenges. 
We will provide the problem formulation and present the experimental settings followed by the key results and analysis of the pilot study.
\subsection{Problem Formulation}

A GUI agent $\mathcal{F}$ is initially trained on an ID dataset $\mathcal{D}_{\text{ID}} = \{(s_i, x_i)\}_{i=1}^k$, consisting of $k$ pairs of screenshots $s_i$ and user instructions $x_i$. $\mathcal{F}$ learns knowledge of operating systems through methods such as SFT and RL on $\mathcal{D}_{\text{ID}}$.

After deployment on the device, $\mathcal{F}$ receives an instruction \( x \) and captures the current device screenshot \( s_t \). $\mathcal{F}$ then constructs a prompt that combines \( s_t \) and \( x \), which is subsequently used to predict an action \( a_t \).
Formally, at each time step \( t \) (\( 0 \leq t \leq T \)), the process can be expressed as:
$a_t = \mathcal{F}(s_t, x)$
where \( s_t \) is the screenshot at time step \( t \), \( x \) is the instruction, and \( a_t \) is the action predicted by the agent.

Then \( a_t \) is executed, leading to a new screenshot \( s_{t+1} \). 
$\mathcal{F}$ evaluates whether the instruction \( x \) has been completed. If \( x \) is not completed, the agent repeats this process until either \( x \) is completed or the maximum step limit \( T \) is reached.
Throughout the execution process, it is possible that some pairs of screenshots and instructions \( (s_t, x) \) may deviate significantly from the distribution of \( \mathcal{D}_{\text{ID}} \). Such pairs are classified as OOD samples. When these OOD samples are encountered, the agent’s action predictions are prone to errors, which can lead to undesirable execution results.

The objective of OOD detection for GUI agents, therefore, is to identify whether the pair \( (s_t, x) \) at each time step \( t \) deviates from the distribution of \( \mathcal{D}_{\text{ID}} \). If OOD is detected, the agent should immediately terminate the execution of the instruction and alert the user; otherwise, it continues with the execution. To formally define the OOD detection mechanism, we introduce the following OOD detection function \( f_{\text{OOD}} \), which operates on each pair \( (s_t, x) \).
If $(s_t, x) \text{ is OOD}$, \( f_{\text{OOD}}(s_t, x) \) returns a value of 1. 
Otherwise, the agent will predict the appropriate action \( a_t \) and proceed to execute it. 

\subsection{Pilot Study with Popular OOD Detection Methods}
Existing popular OOD detection methods~\cite{malinin2018predictive, yang2024generalized} can be broadly categorized into two main types: embedding-based methods~\cite{lee2018simple} and uncertainty-based methods~\cite{gal2016dropout}.
Based on existing research~\cite{dong2024multiood,lu2024recent} in OOD detection for (M)LLMs, we used three embedding-based methods and two uncertainty-based methods for experimentation.

We evaluated the methods used as baselines across eight datasets spanning three platforms. 
Following standard OOD detection method evaluation criteria~\cite{luan2021out,cui2022out}, we plotted the ROC curve and reported the area under the receiver operating characteristic curve (AUROC)~\cite{metz1978basic} and false positive rate at 95\% true positive rate (FPR95).

The pilot experiment results are shown in Table~\ref{pilot}. 
The AUROC metric reflects the separability between OOD and ID datasets achieved by different methods. 
Even the best baseline performances only range from 76.14\% to 89.77\%, barely reaching the acceptable level expected for OOD detection. 
The FPR95 metric measures the false positive rate when the true positive rate reaches 95\%. Pilot experiments show that even the best baselines result in an FPR95 of 31.75\% to 48.92\%, indicating that while maintaining 95\% task success for the GUI agent, 31.75\% to 48.92\% of OOD tasks would be mistakenly executed, posing significant risks.

\begin{figure}[t]
    \centering
    \includegraphics[width=0.46\textwidth]{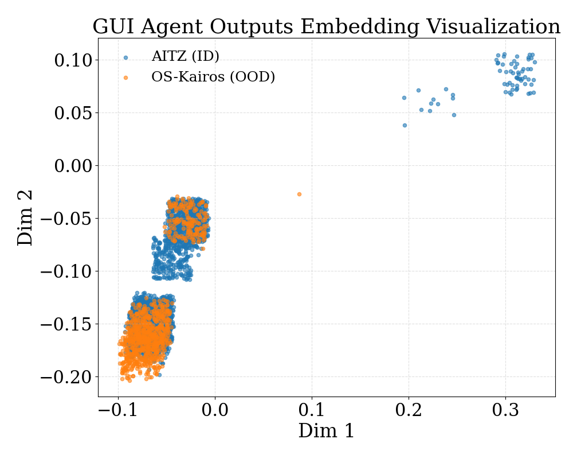}
    \caption{The output embeddings of the OS-Kairos and AITZ are visualized. Most of the samples are confused.}
    \label{output}
\end{figure}
\subsection{Why do these OOD detection methods perform suboptimally?}
For embedding-based methods, due to the relatively uniform reasoning patterns of GUI agents—TV score, which rely on differences in layer embeddings to infer reasoning path divergence, perform suboptimally. 
Other methods based on single-layer embeddings have achieved better performance; however, since GUI agents tend to lose some information related to input understanding during the reasoning process, there is still room to improve.

For uncertainty-based methods, the separability between the output spaces of ID and OOD samples in the GUI agent domain is extremely low (as shown in Figure~\ref{output}). 
As a result, GUI agents struggle to estimate the uncertainty of their outputs, making uncertainty-based methods nearly ineffective to distinguish between ID and OOD samples.

These OOD detection methods rely on finding a decision boundary by identifying the Youden Index~\cite{fluss2005estimation} after obtaining some scoring metric, that is $\text{Youden Index} = \arg\max_{t} \left( \text{TPR}(t) - \text{FPR}(t) \right)$,
where \( t \) is the threshold, and TPR and FPR represent the true positive rate and false positive rate.

\begin{figure}
  \centering
   \includegraphics[width=0.46\textwidth]{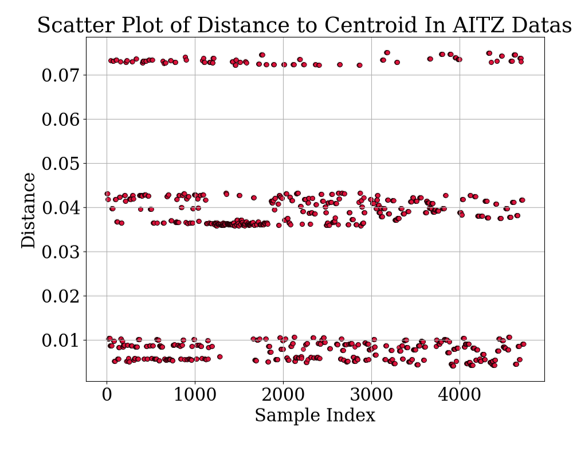}
  \caption{The multimodal embeddings of the AITZ dataset show a clustered distribution pattern around the centroid.}
  \label{discuss1}
\end{figure}
However, popular GUI agents~\cite{wu2024atlas,qin2025ui,xu2024aguvis} have access to diverse training data. In the embedding space, the ID dataset representations for these GUI agents naturally form clusters according to their different data sources. Moreover, we observe that even for datasets originating from a single data source such as AITZ~\cite{zhang2024android}, as shown in Figure~\ref{discuss1}, there can still be noticeable clustering phenomena. We also observe that samples farther from the centroid tend to yield higher success rates.
For such non-linearly separable data, the Youden Index approach becomes inadequate. 

\section{GEM Method}
\label{method}
\begin{table*}[t]
\centering
\begin{tabular}{lcccccccc} 
\toprule
& \multicolumn{4}{c}{AndroidControl} & \multicolumn{4}{c}{OS-Kairos} \\
\cmidrule(lr){2-5} \cmidrule(lr){6-9}
\textbf{Method} & Acc.(\%)$\uparrow$ & Prec.(\%)$\uparrow$ & Rec.(\%)$\uparrow$ & F1(\%)$\uparrow$ & Acc.(\%)$\uparrow$ & Prec.(\%)$\uparrow$ & Rec.(\%)$\uparrow$ & F1(\%)$\uparrow$ \\ 
\midrule
TV score          & 55.37 & 42.48 & 68.95 & 52.57 & 77.36 & 90.20 & 82.20 & 86.01 \\
Top-k confidence  & 68.29 & 77.30 & 71.58 & 74.33 & 63.71 & 26.04 & 74.47 & 38.59 \\
Output entropy    & 68.29 & 55.13 & 62.43 & 58.55 & 63.71 & 93.05 & 61.77 & 74.25 \\ 
Last layer embed. & 70.08 & 69.02 & 96.77 & 80.57 & 32.66 & 17.45 & 91.10 & 29.29 \\ 
Best layer embed. & 78.20 & 77.02 & 94.07 & 84.69 & 74.35 & 33.29 & 67.33 & 44.56 \\ 
\midrule
\textbf{GEM (ours)} & \textbf{99.39} & \textbf{98.33} & \textbf{100.0} & \textbf{99.16} & \textbf{100.0} & \textbf{100.0} & \textbf{100.0} & \textbf{100.0} \\ 
\midrule

& \multicolumn{4}{c}{Meta-GUI} & \multicolumn{4}{c}{ScreenSpot-Mobile} \\
\cmidrule(lr){2-5} \cmidrule(lr){6-9}
\textbf{Method} & Acc.(\%)$\uparrow$ & Prec.(\%)$\uparrow$ & Rec.(\%)$\uparrow$ & F1(\%)$\uparrow$ & Acc.(\%)$\uparrow$ & Prec.(\%)$\uparrow$ & Rec.(\%)$\uparrow$ & F1(\%)$\uparrow$ \\ 
\midrule
TV score          & 68.56 & 71.91 & 91.51 & 80.53 & 65.54 & 92.01 & 67.76 & 78.04 \\
Top-k confidence  & 54.60 & 34.54 & 63.60 & 44.77 & 52.93 & 11.91 & 60.96 & 19.92 \\ 
Output entropy    & 54.60 & 77.46 & 50.93 & 61.46 & 52.93 & 92.62 & 52.07 & 66.67 \\
Last layer embed. & 59.38 & 39.47 & 75.72 & 51.89 & 30.71 & 12.05 & 98.61 & 21.47 \\ 
Best layer embed. & 76.59 & 55.50 & 96.31 & 70.42 & 62.72 & 19.60 & 92.83 & 32.36 \\ 
\midrule
\textbf{GEM (ours)} & \textbf{100.0} & \textbf{100.0} & \textbf{100.0} & \textbf{100.0} & \textbf{100.0} & \textbf{100.0} & \textbf{100.0} & \textbf{100.0} \\ 
\midrule
& \multicolumn{4}{c}{Omniact-Desktop} & \multicolumn{4}{c}{ScreenSpot-Desktop} \\
\cmidrule(lr){2-5} \cmidrule(lr){6-9}
\textbf{Method} & Acc.(\%)$\uparrow$ & Prec.(\%)$\uparrow$ & Rec.(\%)$\uparrow$ & F1(\%)$\uparrow$ & Acc.(\%)$\uparrow$ & Prec.(\%)$\uparrow$ & Rec.(\%)$\uparrow$ & F1(\%)$\uparrow$ \\ 
\midrule
TV score          & 47.50 & 29.04 & 82.36 & 42.94 & 23.92 & 7.66  & 95.21 & 14.18 \\
Top-k confidence  & 44.54 & 84.05 & 33.36 & 47.76 & 31.00 & 95.43 & 27.43 & 42.62 \\
Output entropy    & 44.54 & 27.47 & 79.95 & 40.88 & 31.00 & 7.35  & 81.44 & 13.49 \\
Last layer embed. & 60.05 & 35.80 & 83.84 & 50.17 & 46.54 & 10.05 & 89.22 & 18.06 \\ 
Best layer embed. & \textbf{91.15} & 83.73 & 78.34 & 80.94 & 43.46 & 10.18 & 96.71 & 18.43 \\ 
\midrule
\textbf{GEM (ours)} & 89.53 & \textbf{87.89} & \textbf{100.0} & \textbf{93.55} & \textbf{96.86} & \textbf{96.74} & \textbf{100.0} & \textbf{98.34} \\ 
\midrule

& \multicolumn{4}{c}{Omniact-Web} & \multicolumn{4}{c}{ScreenSpot-Web} \\
\cmidrule(lr){2-5} \cmidrule(lr){6-9}
\textbf{Method} & Acc.(\%)$\uparrow$ & Prec.(\%)$\uparrow$ & Rec.(\%)$\uparrow$ & F1(\%)$\uparrow$ & Acc.(\%)$\uparrow$ & Prec.(\%)$\uparrow$ & Rec.(\%)$\uparrow$ & F1(\%)$\uparrow$ \\ 
\midrule
TV score          & 48.63 & 13.07 & 72.45 & 22.15 & 45.35 & 11.27 & 79.59 & 19.75 \\
Top-k confidence  & 61.36 & 91.17 & 63.15 & 74.61 & 42.97 & 94.91 & 39.84 & 56.12 \\
Output entropy    & 61.36 & 12.16 & 45.47 & 19.19 & 42.97 & 10.54 & 76.83 & 18.54 \\
Last layer embed. & 61.82 & 18.52 & 81.89 & 30.20 & 47.77 & 12.64 & 87.61 & 22.09 \\ 
Best layer embed. & 86.03 & 40.34 & 80.38 & 53.72 & 74.77 & 24.77 & 97.48 & 39.50 \\ 
\midrule
\textbf{GEM (ours)} & \textbf{100.0} & \textbf{100.0} & \textbf{100.0} & \textbf{100.0} & \textbf{100.0} & \textbf{100.0} & \textbf{100.0} & \textbf{100.0} \\
\bottomrule
\end{tabular}
\caption{Comparison of OOD Detection Results Across Different Datasets.}
\label{main}
\end{table*}
\begin{table*}[t]
  \centering
  {
    \begin{tabular}{>{\centering\arraybackslash}p{3cm}
                    >{\centering\arraybackslash}p{1.9cm}
                    >{\centering\arraybackslash}p{2cm}
                    >{\centering\arraybackslash}p{1.9cm}
                    >{\centering\arraybackslash}p{2cm}
                    >{\centering\arraybackslash}p{1.9cm}
                    >{\centering\arraybackslash}p{2cm}}
        \toprule
        \multirow{2}{*}{Model} & \multicolumn{2}{c}{Smartphone} & \multicolumn{2}{c}{Computer} & \multicolumn{2}{c}{Web browser} \\
        \cmidrule(lr){2-3} \cmidrule(lr){4-5} \cmidrule(lr){6-7}
        & Accuracy(\%)$\uparrow$ & F1 Score(\%)$\uparrow$ & Accuracy(\%)$\uparrow$ & F1 Score(\%)$\uparrow$ & Accuracy(\%)$\uparrow$ & F1 Score(\%)$\uparrow$ \\
        \midrule
        UI-TARS-7B & 97.94 & 96.55 & 83.22 & 89.58 & 98.28 & 98.97 \\
        Qwen2-VL-2B & 98.64 & 97.68 & 88.41 & 92.56 & \textbf{100.0} & \textbf{100.0} \\
        Qwen2-VL-7B & 99.51 & 99.16 & 87.63 & 92.10 & \textbf{100.0} & \textbf{100.0} \\
        Qwen2.5-VL-3B & 93.77 & 90.22 & 88.44 & 92.58 & \textbf{100.0} & \textbf{100.0} \\
        Qwen2.5-VL-7B & \textbf{99.78} & \textbf{99.61} & \textbf{89.97} & \textbf{93.50} & \textbf{100.0} & \textbf{100.0} \\
        OS-Atlas-Base-7B & 33.08 & 46.19 & 74.29 & 84.87 & 83.02 & 90.72 \\
        OS-Atlas-Pro-7B & 33.08 & 46.19 & 74.29 & 84.87 & 83.02 & 90.72 \\
        LLaVA1.5 & 28.83 & 44.66 & 72.44 & 83.96 & 83.11 & 90.77 \\
        Blip + BERT & 30.04 & 45.06 & 77.81 & 86.66 & 88.38 & 93.45 \\
        \bottomrule
    \end{tabular}
  }
\caption{The performance of GEM with different encoder structures.}
\label{generalization}
\end{table*}

Pilot experiments show that popular OOD detection methods perform suboptimally in the GUI agent domain. However, we observe that the input embeddings generated by GUI agents naturally lend themselves to modeling the ID representation of $\mathcal{D}_{\text{ID}}$.

Given a GUI agent $\mathcal{F}$ and an ID dataset $\mathcal{D}_{\text{ID}} = \{(s_i, x_i)\}_{i=1}^k$, we first obtain an encoder layer $l_e$ from $\mathcal{F}$. The encoder $l_e$ maps each input pair $(s_i, x_i)$ to an embedding vector $e_i \in {R}^n$.
Thus, we construct an ID embedding dataset $\mathcal{D}_{\text{embedding}} = \{e_i\}_{i=1}^k$. Each $e_i$ is a point in the $n$-dimensional embedding space. 

To model this distribution, we first compute the centroid $\mu$ of $\mathcal{D}_{\text{embedding}}$: $\mu = \frac{1}{k} \sum_{i=1}^{k} e_i$.
Next, we calculate the Euclidean distance between each embedding $e_i$ and the centroid $\mu$: $d_i = \lVert e_i - \mu \rVert_2, \quad i = 1, \ldots, k$,
resulting in a distance dataset $\mathcal{D}_{\text{distance}} = \{d_i\}_{i=1}^k$.

The distribution of $\mathcal{D}_{\text{distance}}$ is typically multi-centered and may contain multiple modes.
Therefore, instead of fitting a simple Gaussian or applying heuristic thresholds, we model it using a GMM. Specifically, we assume that the distances are generated from a mixture of $m$ univariate Gaussian components. The GMM models the probability density function as:
$p(d) = \sum_{j=1}^{m} \pi_j \, \mathcal{N}(d \mid \mu_j, \sigma_j^2)$,
$\pi_j$ is the mixing coefficient of the $j$-th component, satisfying $\sum_{j=1}^{m} \pi_j = 1$ and $\pi_j \geq 0$, $\mathcal{N}(d \mid \mu_j, \sigma_j^2)$ denotes the density of a univariate Gaussian:
\begin{equation}
\mathcal{N}(d \mid \mu_j, \sigma_j^2) = \frac{1}{\sqrt{2\pi\sigma_j^2}} \exp\left( -\frac{(d - \mu_j)^2}{2\sigma_j^2} \right).
\end{equation}

Given the dataset $\mathcal{D}_{\text{distance}}$, the log-likelihood of the data under the GMM is:
\begin{equation}
\log \mathcal{L}_m = \sum_{i=1}^{k} \log \left( \sum_{j=1}^{m} \pi_j \, \mathcal{N}(d_i \mid \mu_j, \sigma_j^2) \right).
\end{equation}

The parameters $\{\pi_j, \mu_j, \sigma_j^2\}$ are estimated by maximizing the log-likelihood $\log \mathcal{L}_m$ via the expectation-maximization (EM) algorithm. Each iteration of EM consists of E-step and M-step.
E-step estimate the posterior probability that sample $d_i$ belongs to component $j$:
\begin{equation}
\gamma_{ij} = \frac{ \pi_j \, \mathcal{N}(d_i \mid \mu_j, \sigma_j^2) }{ \sum_{l=1}^{m} \pi_l \, \mathcal{N}(d_i \mid \mu_l, \sigma_l^2) }.
\end{equation}
M-step update the parameters:
\begin{equation}
\pi_j^{\text{new}} \!\!= \frac{1}{k} \sum_{i=1}^{k} \gamma_{ij},~\mu_j^{\text{new}} \!\!= \frac{ \sum_{i=1}^{k} \gamma_{ij} d_i }{ \sum_{i=1}^{k} \gamma_{ij} },
\end{equation}

\begin{equation}
\sigma_j^{2\,\text{new}} \!\!= \frac{ \sum_{i=1}^{k} \gamma_{ij} (d_i - \mu_j^{\text{new}})^2 }{ \sum_{i=1}^{k} \gamma_{ij} }.
\end{equation}
The E-step and M-step are alternated until convergence to a local maximum of the likelihood.

To determine the optimal number of components $m$, we employ the BIC, defined as:
\begin{equation}
\text{BIC}(m) = -2 \log \mathcal{L}_m + m \log k,
\end{equation}

The optimal number of components $m^*$ is thus selected as $m^* = \arg\min_m \text{BIC}(m)$.
Once the GMM is fitted with $m^*$ components, each Gaussian component $\mathcal{N}(\mu_j, \sigma_j^2)$ provides a center $\mu_j$ and a standard deviation $\sigma_j$. We define the ID boundary as the interval $[\mu_j - n\sigma_j, \mu_j + n\sigma_j]$.

At inference time, given a new input pair $(s_t, x)$, the GUI agent computes the embedding $e_t = l_e(s_t, x)$ and its distance to the centroid $d_t = \lVert e_t - \mu \rVert_2$.

To determine whether $(s_t, x)$ is ID, we check whether $d_t$ falls within any of the ID boundaries defined by the fitted GMM components. Formally, the OOD detection function $f_{\text{OOD}}(s_t, x)$ is given by:
\begin{equation}
f_{\text{OOD}}(s_t, x) =
\begin{cases}
0, & \text{if } \exists j \text{ s.t. } d_t \in [\mu_j - n\sigma_j, \mu_j + n\sigma_j], \\
1, & \text{otherwise}.
\end{cases}
\end{equation}

If $f_{\text{OOD}}(s_t, x) = 1$, the input is classified as OOD, and the agent immediately terminates execution and notifies the user. Otherwise, the agent proceeds with its normal action prediction and execution.

\section{Experiments}
\label{experiments}
In this section, we first introduce the experimental setup, followed by a presentation and analysis of the performance of GEM on OOD detection for GUI agents.
\subsection{Experiments Setup}
\label{setup}
\subsubsection{Datasets.} 
For the ID dataset, we use the AITZ~\cite{zhang2024android} dataset, which contains GUI agent data covering more than 70 Android app scenarios. 
For the OOD datasets, we select eight GUI agent datasets spanning three platforms: smartphone, computer, and web browser. 
The smartphone platform includes AndroidControl~\cite{li2024effects}, OS-Kairos~\cite{cheng2025kairos}, Meta-GUI~\cite{sun2022meta}, and ScreenSpot-Mobile~\cite{li2025ScreenSpot}, while the computer platform includes Omniact-Desktop~\cite{kapoor2024omniact} and ScreenSpot-Desktop, and the web browser platform includes Omniact-Web and ScreenSpot-Web. 

\begin{figure*}[t]
  \centering
  \includegraphics[width=\textwidth]{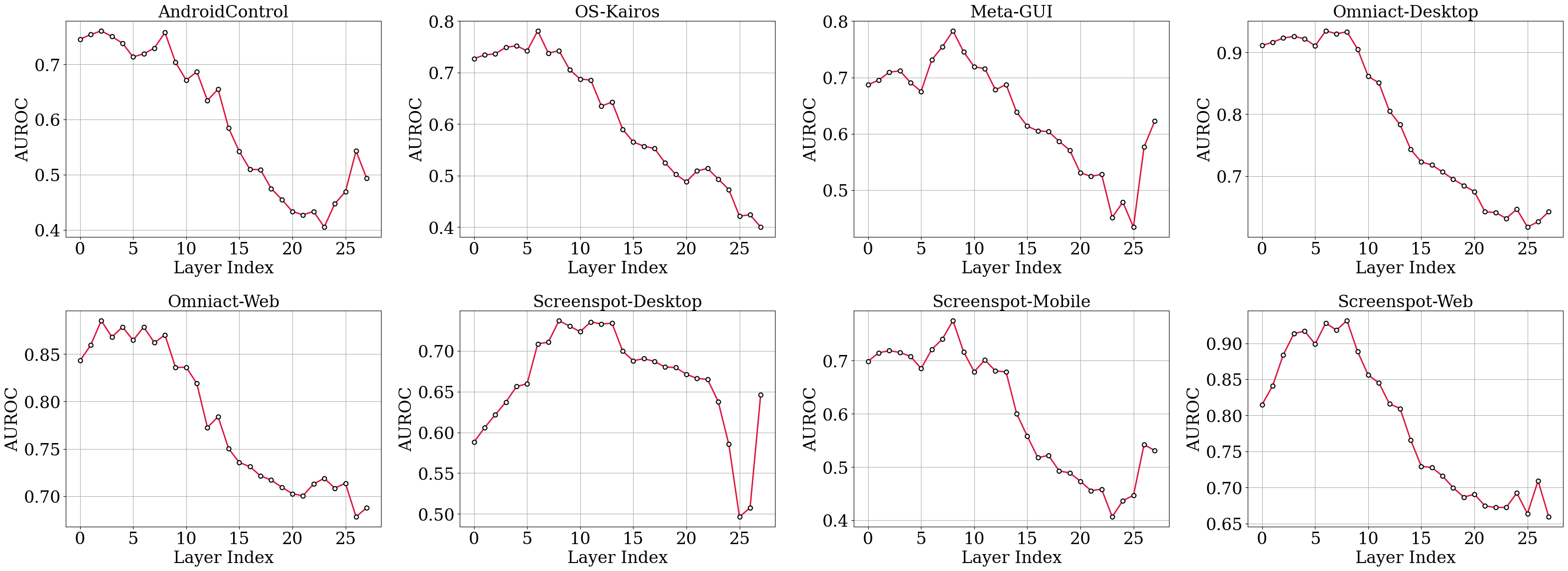}
  \caption{AUROC for OOD detection using embeddings from different layers of the GUI agent.}
  \label{discuss2}
\end{figure*}
\begin{table*}
\centering
\begin{tabular}{p{2cm}p{2cm}p{1.8cm}p{1.8cm}p{1.8cm}p{1.8cm}p{1.8cm}} 
\toprule
Mode      & SCROLL(\%)$\uparrow$ & PRESS(\%)$\uparrow$ & STOP(\%)$\uparrow$  & CLICK(\%)$\uparrow$ & TYPE(\%)$\uparrow$  & TOTAL(\%)$\uparrow$  \\ 
\midrule
Zero-Shot & 19.24  & 32.05 & 0.00  & 30.00 & 44.59 & 26.94  \\
SFT       & 63.11  & 32.60 & 65.35 & 45.14 & 44.59 & 49.38  \\
SFT + GEM   & 63.11$_{0.00\uparrow}$  & 35.00$_{2.40\uparrow}$ & 73.20$_{28.06\uparrow}$ & 55.40$_{10.26\uparrow}$ & 52.59$_{8.00\uparrow}$ & 58.78$_{9.40\uparrow}$  \\
\bottomrule
\end{tabular}
\caption{Analysis of experiments to use GEM to improve the end-to-end results. We report the single-step action accuracy of GUI agents for different action types and the total.}
\label{end2end}
\end{table*}
\subsubsection{Implementation.} Popular GUI Agents~\cite{wu2024atlas, qin2025ui, xu2024aguvis} are developed based on Qwen2-VL-7B. 
To simulate the construction process of a popular GUI Agent, we perform SFT on Qwen2-VL-7B using AITZ train dataset, and then evaluate its OOD detection performance on each sample from OOD datasets and AITZ test dataset (ID). 


\subsection{Main Results}
Table~\ref{main} shows the main results.
GEM consistently outperforms existing methods across almost all datasets. Although on the Omniact-Desktop dataset the accuracy is 1.62\% lower than the best-performing baseline, our method achieves substantial improvements in other metrics. This shows a better balance between rejecting OOD samples and retaining ID samples, which is critical for reliable OOD detection.
On other seven datasets, GEM outperforms all baselines across all evaluation metrics.

Because GUI agents exhibit relatively simple reasoning paths, limiting the effectiveness of methods like the TV score, which rely on modeling diverse reasoning paths. And the complex semantics of multimodal inputs weakens the performance of other embedding-based approaches. Furthermore, due to the semantic similarity in the output space, uncertainty-based OOD detection methods for GUI agents are unreliable.
In contrast, GEM effectively captures subtle high-dimensional differences between ID and OOD data, leading to strong and robust detection.

\section{Further Analysis}
\label{analysis}
In this section, we present several interesting observations and propose a potential method for using GEM to enhance end-to-end tasks in GUI agents.
\subsection{Generalization Experiment}

To demonstrate the generalization ability of GEM, we evaluate its performance across nine different GUI agents or MLLMs. As shown in Table~\ref{generalization}, GEM consistently achieves high accuracy across five models spanning three platforms.
For the results where GEM performs suboptimally, OS-Atlas were exposed to most of the smartphone datasets used in our experiment during their GUI grounding pretraining phase. 
Meanwhile, LLaVA-1.5 and BLIP capture visual information at a coarser granularity. 
This highlights their limitations in handling the complex embedding space specific to GUI agent tasks. 

\subsection{Layer-Level OOD Detection Analysis}
We evaluated the effectiveness of representations extracted from each of the twenty-eight layers of Qwen2-VL-7B for OOD detection using an embedding-based approach.
As shown in Figure~\ref{discuss2}, in most datasets, the AUROC first increases as the layer depth increases, then decreases. However, the AUROC rises again in the last two layers. 
Interestingly, in 4 out of 8 datasets, the best OOD detection results are achieved at the ninth layer. 
This might be because, as the layers get deeper, the importance of specific task-related features increases, while the contribution of general visual or textual features decreases. Around the ninth layer, the GUI agent likely finds a best balance. 
And the GUI agent has developed a kind of confidence estimation in its final output in the final two layeers, allowing it to better distinguish between OOD and ID samples.

\subsection{How to use GEM to improve the end-to-end results}
GEM not only enhances the security of GUI agents but also improves the end-to-end results of GUI agents by combining it with other modules. 
For example, we trained Qwen2-VL-7B using the AITZ dataset and conducted tests using a mixed test set of AITZ and OS-Kairos to simulate a real GUI agent testing environment with multiple knowledge sources. 
When GEM determines that the current sample is an OOD sample, it requests assistance from the cloud-based GPT-4o to execute tasks. 
We report the step-wise success rates for different types of GUI agents as well as the total success rate. 
The experimental results are shown in Table~\ref{end2end}, indicating that GEM can enhance the step-wise success rate of GUI agents by 9.4\%, demonstrating the potential of GEM to improve the end-to-end results.

\subsection{Time Cost of GEM}
Although GEM calculates the distances in the embedding space multiple times, it does not actually incur a burden on the time overhead. We conducted experiments on the time cost of GEM, and the experimental results are shown in Table~\ref{time}. The experiments indicate that GEM only increases training time by 4.9\% and testing time by 6.5\%.

\begin{table}
\centering
\begin{tabular}{lcc} 
\toprule
                           & Training time & Testing time \\ 
\midrule
Original time              & 3.63 h    & 0.77 s   \\
Time with GEM              & 0.18 h    & 0.05 s   \\
Additional time (\%) & 4.9     & 6.5    \\
\bottomrule
\end{tabular}
\caption{Time Comparison of GUI Agent w/ and w/o GEM.}
\label{time}
\end{table}

\section{Conclusion}
\label{conclusion}
We present the first systematic analysis of OOD detection for GUI agents and show that traditional OOD detection methods perform suboptimally in this domain. 
Then we propose GEM, a method that uses input embedding information to fit a GMM to detect OOD samples for GUI agents.
Experiments show that GEM achieves an average accuracy improvement of 23.70\% over the best-performing baseline while only increasing training time by 4.9\% and testing time by 6.5\%.
And we present several interesting observations and propose a potential method for using GEM to enhance end-to-end tasks in GUI agents.

\bibliography{aaai2026}

\newpage
\begin{center}
    \fontsize{16}{20}\selectfont\textbf{Appendix}
\end{center}
\section{Ethics Statement}
All datasets and models used in this work are sourced from the official repositories associated with the original papers, and we strictly follow their respective usage protocols. The datasets remain unmodified, and the models are only subjected to supervised fine-tuning and inference on OOD data. To ensure the safety of model outputs, all generated results have been manually reviewed to prevent any potentially harmful or inappropriate content. As our research focuses exclusively on OOD detection for GUI agents and does not involve sensitive or personal data, we believe our work poses minimal risk of societal harm.
\section{Limitations}
\label{limitations}

Due to the evolving environments of GUI agent domain, although our experiments cover major datasets across three platforms—smartphones, computers, and web browsers—there is still a gap between our setup and the simulation of real-world GUI agent environments.

GEM adopts a GMM to fit the distribution of multimodal input semantics in the ID dataset. Although GEM achieves state-of-the-art performance on mainstream benchmark datasets, its underlying assumption—that multimodal semantic distributions can be effectively modeled using a limited number of Gaussian components—may not hold under highly anomalous GUI data distributions, where the semantic space exhibits irregular or complex patterns that are difficult to approximate with a finite mixture model.
\section{Expanded Analysis}
\subsection{The detailed analysis of clustering phenomenon}
We further evaluate the agent's performance across different clusters on the AITZ dataset. s shown in Figure~\ref{discuss3}, we also observe that the farther a sample is scattered from the centroid, the higher the action success rate of the GUI agent tends to be. Interestingly, although the action match rate slightly drops at mid-range distances, it increases significantly again at far distances. This is because, in the embedding hypersphere, samples that lie farther from the centroid are less likely to be confused with others, making their knowledge more distinctly learnable by the model.
\begin{figure}[htbp]
  \centering
  \includegraphics[width=0.48\textwidth]{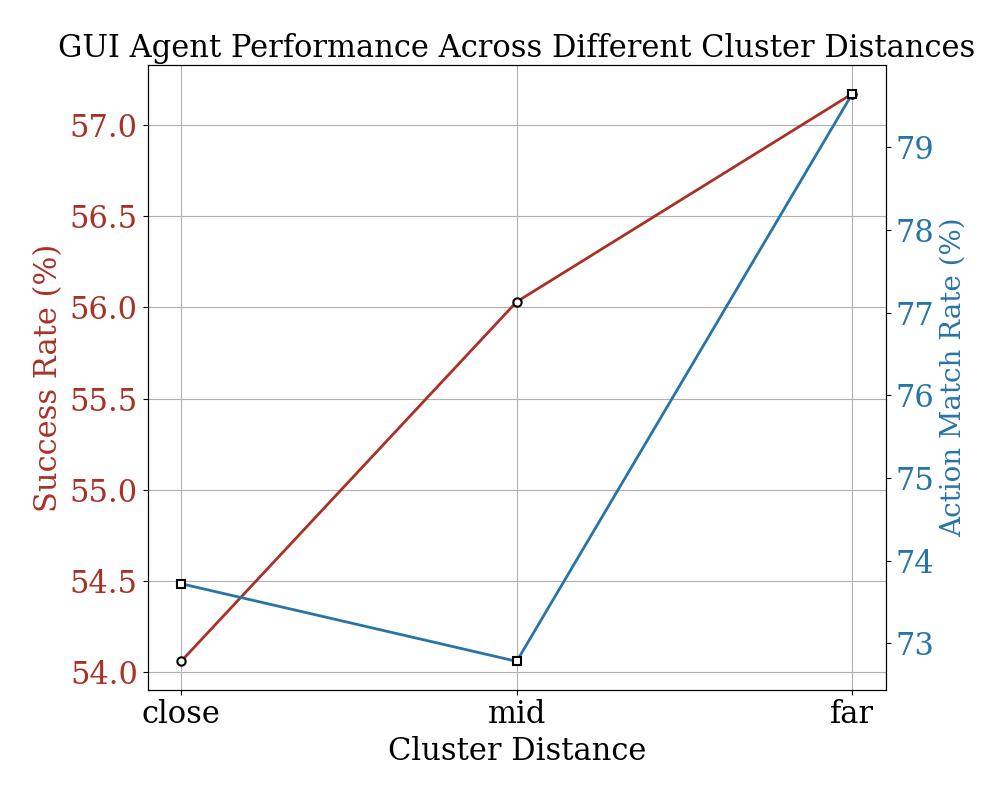}
  \caption{GUI agent performance across different cluster distances.}
  \label{discuss3}
\end{figure}

\subsection{Why does the TV score fail in OOD detection for GUI agents?}
GUI agents operate on terminal devices with very similar thinking logic. The TV score assesses the model's thought process based on the changes in the embeddings of adjacent layers, but it cannot distinguish whether the current scene is OOD or not. As shown in Figure~\ref{discuss3}, the TV score is indistinguishable between OOD samples and ID samples.

\section{Datasets Details}
\label{datasets}
In this section, we provide a detailed description to the datasets used in this paper.

\textbf{AITZ} is the first dataset to annotate Chain-of-Action-Thought (CoAT) for GUI agents, which considers descriptions of prior actions and the current screen, the actions that should be taken, and reasoning about the potential consequences of the selected actions. It contains 18,643 screen-action pairs with CoAT annotations.

\textbf{AndroidControl} is a dataset covering 833 distinct Android applications and containing 15,283 everyday tasks performed within these apps. This dataset is currently the most diverse benchmark for computer-based control, providing a foundation for in-depth model performance analysis.

\textbf{OS-Kairos} is a dataset containing data from 12 different Chinese and English mobile applications across 12 topic categories. The tasks in this dataset were automatically collected by an agent capability probing framework and later curated by humans. It covers a wide range of scenarios encountered in everyday life.

\textbf{Meta-GUI} is a dataset containing dialogues and GUI interaction trajectories. It covers six high-frequency daily-life topics across 11 mobile applications.

\textbf{Omniact} is a dataset consisting of 9.8K pairs of screen images and natural language task descriptions, covering a variety of operating systems and web applications. The dataset includes tasks of varying difficulty levels, ranging from simple single-step actions to complex multi-step operations.

\textbf{ScreenSpot} is a dataset containing high-resolution real screenshots and expert annotations from various domains. This dataset covers data from multiple platforms, including Mac, Windows, Android, iPhone, and web, making it the most platform-diverse GUI dataset.
\begin{figure}[htbp]
\centering
\includegraphics[width=0.48\textwidth]{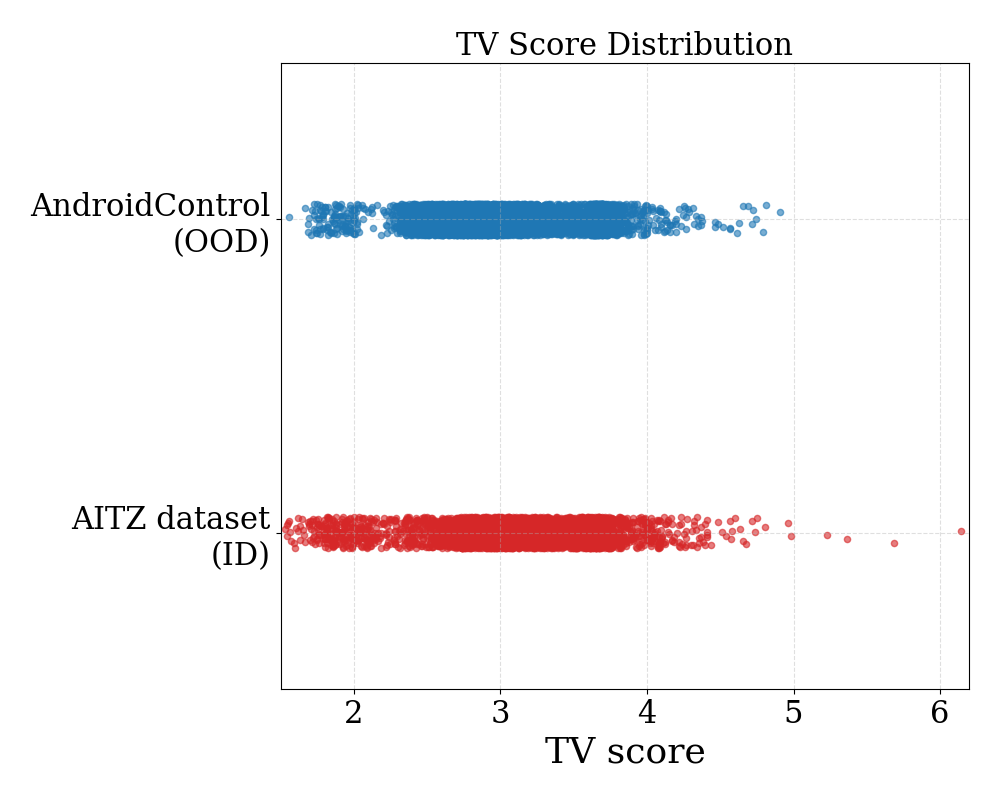}
\caption{The distinguishing effect of the TV score on AndroidControl and AITZ.}
\label{discuss3}
\end{figure}

\section{Main Experiment Details}
\label{main_details}
In this section, we provide more details about our main experiment.

\textbf{Hyperparameter Settings.} All random seeds in our main experiment are set to 42. The number of GMM clusters selected by BIC ranges from 1 to 15. During SFT, we train for 7 epochs with a batch size of 2 and a learning rate of 1.0e-5. The SFT process takes approximately 3.6 hours on 4 A100 GPUs (80GB each).

\begin{figure*}[htbp]
  \centering
  \includegraphics[width=0.8\linewidth]{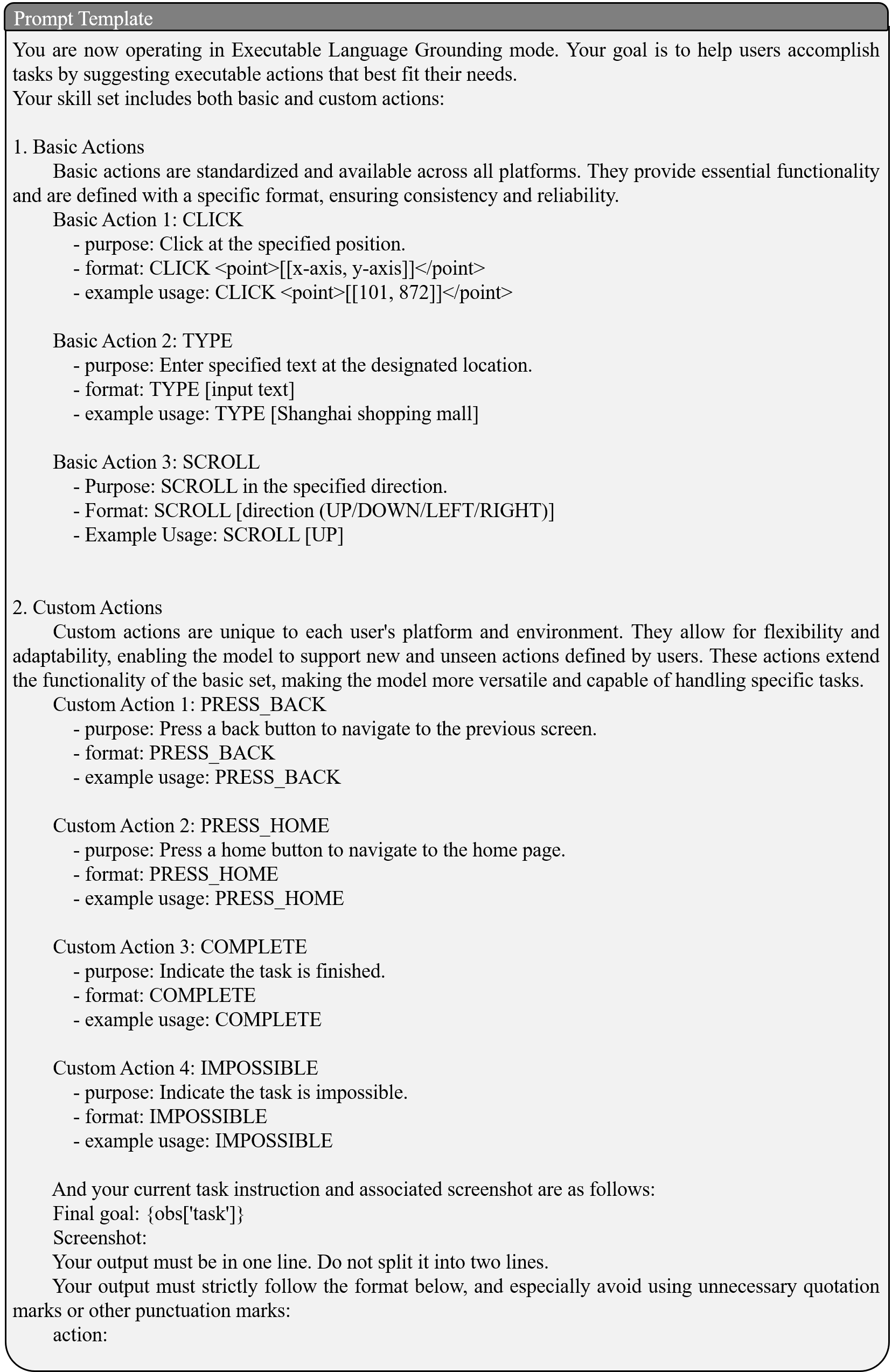}
  \caption{Prompt for GEM and all baselines.}
  \label{fig:prompt}
\end{figure*}

\textbf{Evaluation Metrics Details.} After performing SFT on Qwen2-VL-7B using the AITZ training set, we fit a distance-based GMM on the same training set following the GEM method. Classification boundaries are defined using three standard deviations per cluster. The accuracy, precision, recall, and F1 scores reported in main results are obtained by treating each dataset as the OOD samples and the AITZ test set as the ID samples. The prompt of GEM and all baselines is shown in Figure~\ref{fig:prompt}.

\section{Baseline Details}
\label{baseline}
In this section, we provide a detailed description to the baseline methods used in this paper.

\subsection{TV score}
TV score is an OOD detection method that has been proven effective in the domain of mathematical reasoning with LLMs.
The TV score measures the instability or variation of a test sample’s layer-wise hidden representations with respect to a Gaussian distribution fitted to ID samples. The computation proceeds as follows:
Let $\hat{y}_l^{(i)}$ represent the hidden representations of the $i$-th ID sample at layer $l$, where $l$ ranges from 1 to $L$ and $i$ ranges from 1 to $N$. Let $y_l$ represent the hidden representations of the test sample being evaluated at layer $l$, where $l$ ranges from 1 to $L$.

For each layer $l$, we estimate a Gaussian distribution $\mathcal{G}_l = \mathcal{N}(\mu_l, \Sigma_l)$ by computing the empirical mean $\mu_l$ and covariance $\Sigma_l$ of the ID embeddings at that layer. The out-of-distribution (OOD) score for the evaluated sample at layer $l$ is then defined as the Mahalanobis distance:
\begin{equation}
f(\mathbf{y}_l) = (\mathbf{y}_l - \mu_l)^\top \Sigma_l^{-1} (\mathbf{y}_l - \mu_l).
\label{eq:mahalanobis}
\end{equation}

To assess higher-order fluctuations in the layer-wise representation trajectory, we further define a smoothed $i$-th order differential form of the representation. For each differential order $i = 1, \dots, k$, we compute transformed Gaussian parameters:
\begin{equation}
\mu_l^{(i)} = \sum_{t=0}^{i} (-1)^{i+t} \binom{i}{t} \mu_{l+t}, \quad
\Sigma_l^{(i)} = \sum_{t=0}^{i} \binom{i}{t} \Sigma_{l+t}.
\label{eq:diff-gaussian}
\end{equation}

Using these, we compute the Mahalanobis distance for the $i$-th order differential:
\begin{equation}
f^{(i)}(\mathbf{y}_l) = (\mathbf{y}_l^{(i)} - \mu_l^{(i)})^\top (\Sigma_l^{(i)})^{-1} (\mathbf{y}_l^{(i)} - \mu_l^{(i)}),
\label{eq:diff-mahalanobis}
\end{equation}
where $\mathbf{y}_l^{(i)}$ denotes the $i$-th order smoothed difference of the sample’s layer-wise representations, constructed analogously to the means.

Finally, the TV score at differential order $i$ is obtained by averaging the per-layer scores:
\begin{equation}
\mathrm{TV}_i = \frac{1}{L} \sum_{l=1}^{L} f^{(i)}(\mathbf{y}_l).
\label{eq:tv-score}
\end{equation}

\subsection{Top-k confidence}
Top-$k$ confidence is a likelihood-based OOD detection method that evaluates the probability assigned by the language model to its most confident completions.
Given a test input, the model generates $k$ candidate output sequences, each consisting of a sequence of tokens. Let $P_j = \prod_{t=1}^{T_j} p(y_t^{(j)} \mid y_{<t}^{(j)}, x)$ denote the joint probability of the $j$-th output sequence, where $x$ is the input, $y_t^{(j)}$ is the $t$-th token in the $j$-th output, and $T_j$ is the output length. The Top-$k$ confidence score is then defined as:
\begin{equation}
\mathrm{Top\text{-}k} = \max_{j \in {1, \dots, k}} P_j.
\end{equation}
A lower Top-$k$ confidence score indicates a higher likelihood that the input is out-of-distribution, as the model fails to assign high probability to any of its top candidates.

\subsection{Output entropy}
Output entropy captures the model’s uncertainty over the space of generated sequences and provides a distributional measure of output dispersion.
For the same $k$ candidate sequences used in Top-$k$ confidence, let $P_j$ denote the joint probability of the $j$-th output sequence. The normalized distribution over candidates is given by:
\begin{equation}
\tilde{P}_j = \frac{P_j}{\sum_{i=1}^{k} P_i}.
\end{equation}
The entropy of this distribution is then computed as:
\begin{equation}
\mathrm{Entropy} = -\sum_{j=1}^{k} \tilde{P}_j \log \tilde{P}_j.
\end{equation}
Higher entropy suggests that the model is more uncertain about its output space, and such uncertainty often correlates with inputs being OOD.

\subsection{Last layer embedding}
This method assesses distributional proximity in the feature space of the model’s final layer.
Let $\mathbf{y}_L$ denote the representation of the test sample at the final layer $L$, and let $\{\hat{y}_L^{(i)}\}_{i=1}^{N}$ be the final-layer representations of the $N$ in-distribution (ID) samples. We compute the empirical mean of the ID embeddings as:
\begin{equation}
\mu_L = \frac{1}{N} \sum_{i=1}^{N} \hat{y}_L^{(i)}.
\end{equation}
The OOD score is then defined as the Euclidean distance from the test sample’s final-layer representation to this mean:
\begin{equation}
f(y_L) = |y_L - \mu_L |_2.
\end{equation}
A larger distance indicates that the sample lies further from the ID cluster in representation space, suggesting a higher likelihood of being OOD.

\subsection{Best layer embedding}
Best layer embedding extends the previous approach by selecting the most discriminative layer for OOD detection.
For each layer $l \in \{1, \dots, L\}$, we compute the per-layer representation $\mathbf{y}_l$ of the test sample and the corresponding ID mean $\mu_l$ using the ID samples. The distance is computed similarly as:
\begin{equation}
f_l(y_l) = |y_l - \mu_l |_2.
\end{equation}
To determine the best layer $l^*$, we evaluate the OOD detection performance of each layer using a held-out validation set and select the one achieving the highest AUROC. The final score is then:
\begin{equation}
\mathrm{BestLayerScore} = f_{l^*}(y_{l^*}).
\end{equation}
This method allows flexibility in choosing the most informative representation space for distinguishing OOD inputs.

\section{GEM Algorithm}
\label{algorithm}
In this section, we present the algorithm of our method in Algorithm ~\ref{alg:gmm-ood}.
\begin{algorithm}[htbp]
\caption{GEM Algorithm}
\label{alg:gmm-ood}
\begin{algorithmic}[1]
\Require GUI agent $\mathcal{F}$, encoder layer $l_e$, ID dataset $\mathcal{D}_{\text{ID}} = \{(s_i, x_i)\}_{i=1}^k$
\Ensure OOD detection function $f_{\text{OOD}}$

\State $\mathcal{D}_{\text{embedding}} \gets \emptyset$

\For{$i \gets 1$ \textbf{to} $k$}
    \State $e_i \gets l_e(s_i, x_i)$
    \State $\mathcal{D}_{\text{embedding}} \gets \mathcal{D}_{\text{embedding}} \cup \{e_i\}$
\EndFor

\State $\mu \gets \frac{1}{k} \sum_{i=1}^{k} e_i$

\State $\mathcal{D}_{\text{distance}} \gets \emptyset$

\For{$i \gets 1$ \textbf{to} $k$}
    \State $d_i \gets \lVert e_i - \mu \rVert_2$
    \State $\mathcal{D}_{\text{distance}} \gets \mathcal{D}_{\text{distance}} \cup \{d_i\}$
\EndFor

\For{$m \in \{1, \dots, M\}$}
    \State Initialize GMM parameters $\{\pi_j, \mu_j, \sigma_j^2\}_{j=1}^{m}$
    
    \Repeat
        \For{$i \gets 1$ \textbf{to} $k$}
            \For{$j \gets 1$ \textbf{to} $m$}
                \State $\gamma_{ij} \gets \frac{ \pi_j \, \mathcal{N}(d_i \mid \mu_j, \sigma_j^2) }{ \sum_{l=1}^{m} \pi_l \, \mathcal{N}(d_i \mid \mu_l, \sigma_l^2) }$
            \EndFor
        \EndFor
        
        \For{$j \gets 1$ \textbf{to} $m$}
            \State $\pi_j \gets \frac{1}{k} \sum_{i=1}^{k} \gamma_{ij}$
            \State $\mu_j \gets \frac{ \sum_{i=1}^{k} \gamma_{ij} d_i }{ \sum_{i=1}^{k} \gamma_{ij} }$
            \State $\sigma_j^2 \gets \frac{ \sum_{i=1}^{k} \gamma_{ij} (d_i - \mu_j)^2 }{ \sum_{i=1}^{k} \gamma_{ij} }$
        \EndFor
    \Until {convergence}

    \State $\log \mathcal{L}_m \gets 0$
    \For{$i \gets 1$ \textbf{to} $k$}
        \State $\ell_i \gets \sum_{j=1}^{m} \pi_j \cdot \mathcal{N}(d_i \mid \mu_j, \sigma_j^2)$
        \State $\log \mathcal{L}_m \gets \log \mathcal{L}_m + \log(\ell_i)$
    \EndFor

    \State $\text{BIC}(m) \gets -2 \log \mathcal{L}_m + m \log k$
\EndFor

\State $m^* \gets \arg\min_m \text{BIC}(m)$

\State Fit final GMM with $m^*$ components: $\{(\pi_j, \mu_j, \sigma_j)\}_{j=1}^{m^*}$

\For{$j \gets 1$ \textbf{to} $m^*$}
    \State Define ID interval $I_j = [\mu_j - 3\sigma_j,\ \mu_j + 3\sigma_j]$
\EndFor

\State Define $f_{\text{OOD}}(s, x)$ as:
\begin{equation*}
    f_{\text{OOD}}(s, x) =
    \begin{cases}
        0, & \text{if } \lVert l_e(s, x) - \mu \rVert_2 \in \bigcup_{j=1}^{m^*} I_j \\
        1, & \text{otherwise}
    \end{cases}
\end{equation*}

\Return $f_{\text{OOD}}$
\end{algorithmic}
\end{algorithm}

\section{Ablation Study}
\label{ablation}
In this section, we conduct a ablation study on several adjustable parameters of the GEM algorithm. In all experiments presented in the main text, the number of clusters m for BIC search ranges from 1 to 10, and each GMM cluster defines classification boundaries based on three standard deviations. In this section, we perform an ablation study focusing on the number of clusters m used in BIC search and the granularity of the classification boundary.
\begin{figure*}[t]
  \centering
  \includegraphics[width=\textwidth]{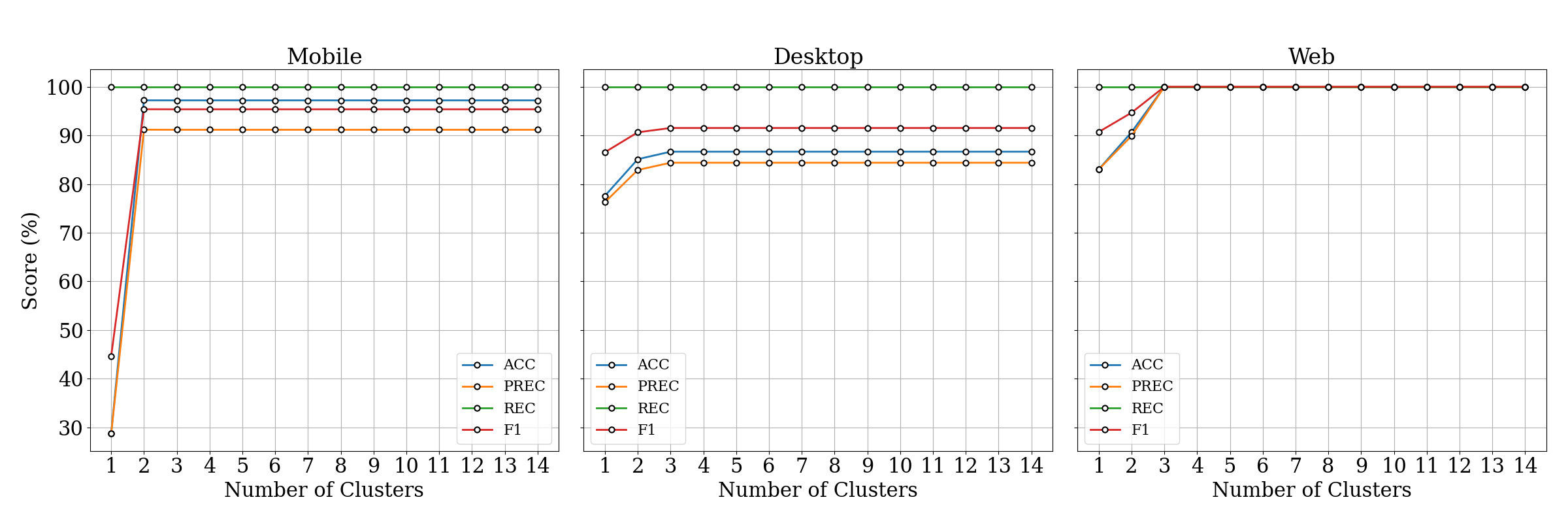}
  \caption{Ablation study on the maximum number of clusters.}
  \label{ablation1}
\end{figure*}
As shown in Figure~\ref{ablation1}, we set the maximum number of clusters for the BIC search from 1 to 15. We then report the performance of GEM on mobile, desktop, and web platforms under different cluster number ranges. We observe that once the number of clusters reaches a certain threshold, further increases have negligible impact on performance, indicating that the selected range likely encompasses the true number of clusters and is therefore reasonable.

\begin{figure*}[t]
  \centering
  \includegraphics[width=\textwidth]{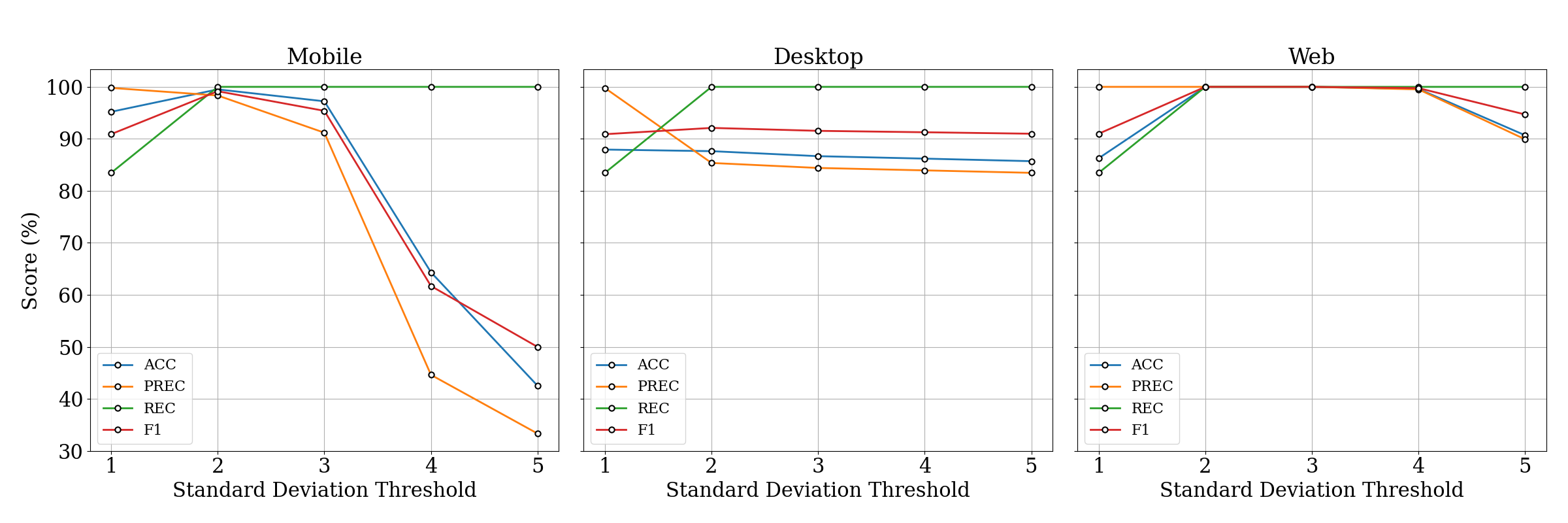}
  \caption{Ablation study on the number of standard deviations as the classification threshold.}
  \label{ablation2}
\end{figure*}

As shown in Figure~\ref{ablation2}, we set different standard deviation thresholds for defining classification boundaries, ranging from 1 to 5 standard deviations. We also report the performance of GEM on mobile, desktop, and web platforms under these settings. The results show that when using 2 standard deviations as the classification boundary, the accuracy reaches its highest on all platforms. However, if the classification boundaries are further widened, several other metrics decline significantly. This is because broader classification boundaries lead to more OOD samples being misclassified as ID samples. Therefore, using 3 standard deviations as the classification boundary is a reasonable choice.

\section{OOD Scenario Examples}
In this section, we show OOD scenario examples classified by internalization-OOD and extrapolation-OOD scenarios in detail.

\begin{figure*}[t]
  \centering
  \includegraphics[width=\textwidth]{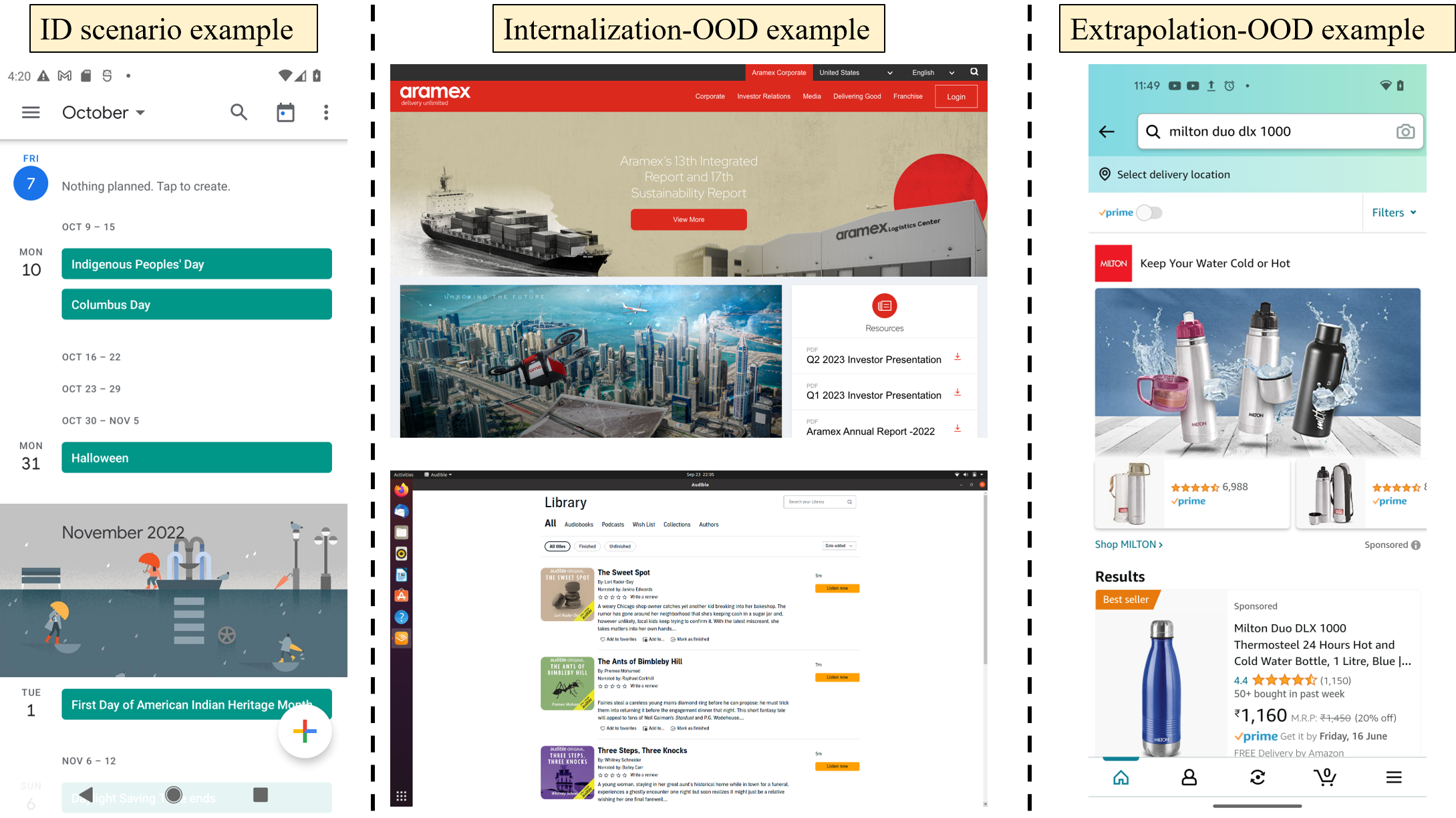}
  \caption{The examples of internalization-OOD and extrapolation-OOD scenarios.}
  \label{scenario}
\end{figure*}

As shown in Figure~\ref{scenario}, for a GUI agent trained on the ID dataset AITZ, it has only acquired knowledge related to operating the smartphone platform. Therefore, for this GUI agent, tasks involving the computer and web browser platforms fall under internalization-OOD scenarios. Since the AITZ dataset does not contain operational knowledge about the Amazon application, any task involving Amazon would constitute an extrapolation-OOD scenario for this GUI agent. In these OOD scenarios, the agent's behavior is less reliable and may pose unnecessary risks.

\section{Why Qwen2-VL-2B Performs Better than Qwen2.5-VL-3B for GEM?}
The performance discrepancy between Qwen2-VL-2B and Qwen2.5-VL-3B in our encoder comparison experiments stems from fundamental architectural differences in their vision encoders. While both models belong to the Qwen-VL family, their encoders are not identical:

Qwen2-VL-2B employs a standard ViT (Vision Transformer) architecture.

Qwen2.5-VL-3B uses a custom window-based attention encoder, where images are split into non-overlapping windows (112×112, equivalent to 8×8 patches). Each window computes self-attention locally, with no cross-window interaction initially, followed by a late-stage full-attention block.

Since GEM relies solely on the encoder component, the model’s total parameter count (e.g., 2B vs. 3B) does not directly predict its effectiveness in our framework. The superior performance of Qwen2-VL-2B suggests that ViT’s global attention mechanism is better suited for GEM’s objectives compared to Qwen2.5-VL-3B’s hybrid window/full-attention design.

\end{document}